# Governance-Aware Federated Learning for Multi-Agent AI Systems: A Policy-Driven Framework for Trustworthy Collaborative AI

Kiranmayee Janardhan[1[0000-0001-6003-4155]], Karthick S[1[0009-0008-8536-2129]], Madhusmita Panda[1], Bharathi Vokkaliga Ganesh[1] and Syed Ahmed[1]

[1] Responsible AI Office, Infosys Limited, Infosys Avenue, 3rd Cross, Electronic City, Hosur Road, Bengaluru – 560100, Karnataka, India
kiranmayee.j@infosys.com

**Abstract.** Federated Learning (FL) enables collaborative model training without exchanging raw data, but conventional FL provides limited runtime governance, evidence verification, fairness assessment, explainability monitoring, and auditability. This paper proposes Governance-Aware Federated Learning (GAFL), which combines independent governance-evidence verification, temporal trust, fairness and explanation assessment, anomaly detection, policy enforcement, normalized aggregation, and tamper-evident audit logging. Client-reported governance values are authenticated, bound to their submitted model updates, and independently recomputed before aggregation. A cross-silo prototype with 10 participants was evaluated on the Adult Income dataset under honest, benign non-IID, biased-data, malicious-update, non-compliant, falsified-evidence, and mixed-threat scenarios. Four methods were compared across five random seeds: FedAvg, FedAvg with anomaly filtering, GAFL without independent verification, and complete GAFL. Complete GAFL intervened on all affected participants at the final round across the evaluated risk scenarios, including identifying 100% of falsified-evidence participants, whereas GAFL without verification identified none of the falsified-evidence participants. Final macro-F1 ranged from 0.7520 to 0.7733 across GAFL scenarios. Complete GAFL required 173.67 ± 26.94 ms of governance processing per round, compared with 3.82 ± 0.79 ms for FedAvg. A strict demographic-parity threshold also caused frequent interventions for honest participants, demonstrating a significant fairness-policy and utility trade-off.

## 1 Introduction

The rapid adoption of distributed artificial intelligence systems has accelerated the need for collaborative machine learning architectures capable of preserving privacy while enabling cross-organizational model development. Federated Learning (FL) enables decentralized entities to collaboratively train machine learning models without directly sharing raw data, thereby improving privacy preservation and regulatory compliance across distributed ecosystems [1]. Traditional federated learning systems primarily focus on distributed optimization and privacy-preserving model aggregation [1][3]. However, enterprise deployment of federated AI introduces additional challenges involving governance, accountability, fairness validation, explainability, trust assessment, anomaly detection, and runtime policy enforcement. These concerns become increasingly important when collaborative environments involve semi-trusted participants operating across organizational boundaries. Existing AI frameworks such as the NIST AI Risk Management Framework and ISO/IEC 42001 emphasize governance, transparency, accountability, fairness, and risk management for AI systems [6][7]. However, these governance principles are not deeply integrated into federated learning workflows, where model updates are typically aggregated without continuous governance validation.

This paper proposes a Governance-Aware Federated Learning (GAFL) framework that embeds runtime governance directly into collaborative learning workflows through a dedicated governance control plane responsible for participant validation, trust evaluation, anomaly detection, fairness and explainability assessment, policy enforcement, and audit coordination. Unlike conventional FedAvg, GAFL conditions each participant's aggregation influence on independently verified governance behaviour and separates predictive-model training from governance evaluation and decision-making. The framework is evaluated against FedAvg, anomaly-filtered FedAvg, and a GAFL ablation without independent evidence verification. In this work, "multi-agent AI systems" denotes coordinated participant and governance agents operating around federated machine-learning models; it does not denote autonomous generative or large-language-model agents.

This work contributes: (1) a round-level governance control plane; (2) non-negative normalized governance-aware aggregation; (3) bounded temporal trust based on verified metrics; (4) independent verification of client-reported governance evidence; and (5) empirical evaluation under honest, non-IID, biased, malicious, non-compliant, falsified-evidence, and mixed-threat conditions.

## 2 Related Work

Federated Averaging enables decentralized collaborative training, while subsequent work has addressed communication, privacy, secure aggregation, heterogeneity, and adversarial robustness [1–3,8,9]. Explainability methods such as SHAP and LIME provide feature-level interpretations [4,5], while demographic parity and related metrics support centralized fairness assessment [10]. NIST AI RMF and ISO/IEC 42001 provide organizational governance guidance but do not define round-level federated governance workflows [6,7]. Trust-weighted and Byzantine-robust methods address related but narrower problems. FLTrust derives client trust from directional similarity to a server update trained on a trusted root dataset [14]. q-FFL modifies the federated optimization objective to improve the uniformity of predictive performance across devices [15]. Krum selects an update using distances among participant updates [16], while coordinate-wise median and trimmed-mean methods provide robust coordinate-level aggregation [17]. GAFL does not claim to replace these optimization or Byzantine-robust aggregation methods. Its contribution is the integration of independently verified governance evidence, multi-metric temporal trust, protected-group fairness and explanation assessment, policy enforcement, and tamper-evident decision records within a round-level governance architecture.

The proposed Governance-Aware Federated Learning framework extends prior work by integrating governance validation mechanisms directly into distributed machine learning orchestration and aggregation workflows. The framework combines runtime governance enforcement, trust-aware aggregation, fairness validation, explainability verification, anomaly monitoring, and auditability within a unified collaborative learning architecture.

## 3 Governance-Aware Federated Learning Framework

### 3.1 System Architecture

The proposed Governance-Aware Federated Learning (GAFL) framework consists of five primary architectural components: cross-silo participant nodes, an Independent Governance Evidence Verifier, a Governance Orchestration Layer functioning as the AI Governance Control Plane, a Governance-Aware Aggregation Server, and a Tamper-Evident Audit and Runtime Monitoring Layer. The participant nodes perform local model training and generate preliminary governance evidence. The Independent Governance Evidence Verifier authenticates each participant, binds the evidence to the submitted model update, and independently recomputes selected governance measurements using a trusted governance-validation dataset. The Governance Orchestration Layer evaluates verified policy-compliance, fairness, explanation-consistency, anomaly-risk, and historical trust information before assigning an accept, down-weight, or quarantine decision. Eligible participant updates are then processed by the Governance-Aware Aggregation Server using non-negative normalized aggregation weights. Governance decisions, verified scores, trust transitions, and aggregation outcomes are recorded by the audit and runtime monitoring layer. The complete architecture is illustrated in Fig. 1.

Each participant trains locally and submits its update with signed preliminary fairness, explanation, compliance, and integrity evidence. Because participants are semi-trusted, these values remain unverified until the Independent Governance Evidence Verifier authenticates the evidence, binds it to the update hash, and independently recomputes the required metrics. The Governance Control Plane then combines verified compliance, fairness, explanation, anomaly, and historical trust information to assign accept, down-weight, or quarantine. Eligible updates proceed to normalized aggregation, while decisions and weights are recorded in the tamper-evident audit layer.

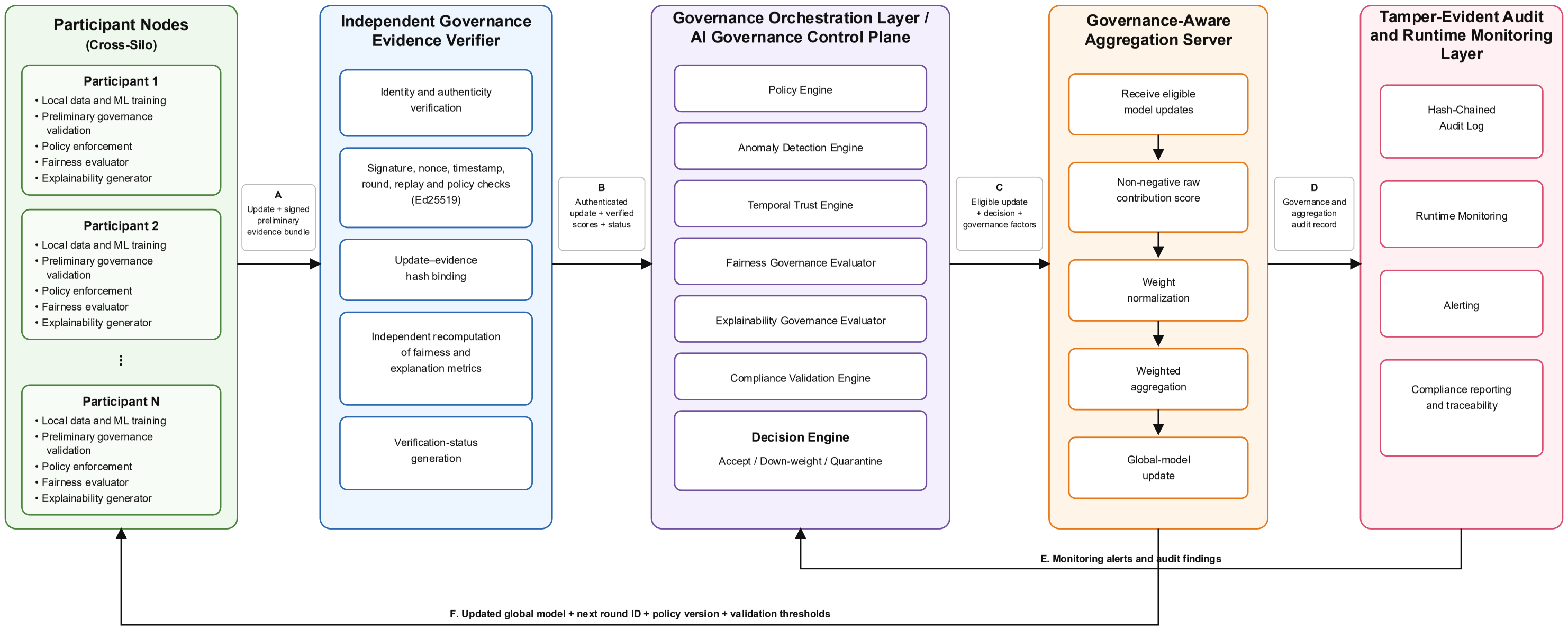

**Fig. 1.** GAFL architecture with independent governance-evidence verification and tamper-evident audit monitoring.

The Temporal Trust Engine maintains a bounded historical trust score $T_i^{(t)} \in [0, 1]$ for participant $i$at federated round $t$. Trust is calculated only from governance measurements verified by the Independent Governance Evidence Verifier. The verified policy-compliance score $C_i^{(t)}$, fairness score $F_i^{(t)}$, and explainability -consistency score $E_i^{(t)}$are normalized to the interval [0, 1], where larger values indicate better governance performance. The anomaly-risk score $A_i^{(t)} \in [0,1]$uses the opposite interpretation, where a larger value indicates greater update risk. The instantaneous governance-quality score and temporal trust update are defined as:

$$Q_i^{(t)} = \lambda_C C_i^{(t)} + \lambda_F F_i^{(t)} + \lambda_E E_i^{(t)} + \lambda_A \left(1 - A_i^{(t)}\right),$$

$$T_i^{(t)} = \rho T_i^{(t-1)} + (1-\rho) Q_i^{(t)}. \quad (1)$$

The governance coefficients satisfy

$$\lambda_C, \lambda_F, \lambda_E, \lambda_A \geq 0 \text{ and}$$
$$\lambda_C + \lambda_F + \lambda_E + \lambda_A = 1.$$

The parameter $\rho \in [0,1)$ controls the influence of historical behaviour: larger values produce slower trust changes, whereas smaller values give greater importance to recent governance outcomes. New participants are initialized with the neutral value $T_i^{(0)} = 0.5$. Mandatory identity, integrity, evidence, or policy failures trigger quarantine rather than being compensated for by stronger scores on other governance dimensions; for such a round, $Q_i^{(t)}$is set to zero. Participants may gradually recover trust through subsequent verified compliant behaviour.

The Explainability Governance Evaluator assesses whether a participant model satisfies the explainability requirements defined for the current federated round. Each participant generates preliminary feature-attribution evidence using closed-form linear SHAP and includes the normalized feature-attribution vector in its signed governance-evidence bundle [4]. Because the participant may be malicious or non-compliant, the locally reported explainability value is not used directly in trust calculation or aggregation. The Independent Governance Evidence Verifier independently recomputes the linear SHAP attribution vector on a fixed trusted governance-validation sample and derives a verified explanation-consistency score.

Let $\mathbf{e}_i^{(t)}$denote the normalized feature-attribution vector produced for participant $i$during round $t$, and let $\mathbf{e}_i^{(t-1)}$denote the verified attribution vector from the preceding evaluated round. The explanation-consistency score is defined as:

$$E_i^{(t)} = \frac{1 + \rho_s\left(e_i^{(t)}, e_i^{(t-1)}\right)}{2} \quad (2)$$

where $\rho_s$ denotes Spearman rank correlation. This transformation maps the score to the interval $[0, 1]$, where larger values indicate more stable feature-attribution rankings across rounds. For a participant's first evaluated round, $E_i^{(0)}$ is calculated by comparing the participant explanation with the explanation produced by the current global model on the same governance-validation sample. The explanation-consistency score measures explanation stability and consistency; it does not establish causal correctness or guarantee that the model prediction itself is correct.

The Fairness Governance Evaluator assesses the participant model using the protected attribute and fairness metric defined for the experimental task. For the Adult Income experiment, demographic-parity difference was used as the primary fairness measure, with sex as the protected attribute [10]. The participant may calculate a preliminary fairness value locally, but this value is treated as an unverified claim. The Independent Governance Evidence Verifier independently recomputes the fairness metric using the trusted governance-validation dataset.

For participant $i$ in round $t$, demographic-parity difference, $D_i^{(t)}$, is defined as:

$$D_i^{(t)} = \left|Pr\left(\hat{Y} = 1 \mid S = 0\right) - Pr\left(\hat{Y} = 1 \mid S = 1\right)\right|$$

and the normalized fairness score is:

$$F_i^{(t)} = 1 - D_i^{(t)} \quad (3)$$

Here, $S$ denotes the protected-group variable and $\hat{Y} = 1$ denotes the favourable predicted outcome. Because $D_i^{(t)} \in [0,1]$, the resulting fairness score also satisfies $F_i^{(t)} \in [0,1]$. A larger fairness score indicates a smaller measured difference in positive prediction rates between the protected groups. The governance policy specifies a demographic-parity threshold $\tau_D$. An update violates the fairness policy when $D_i^{(t)} > \tau_D$. The primary experiments used $\tau_D = 0.10$ while the sensitivity analysis used $\tau_D = 0.15$. Demographic parity is used only as the primary metric for the evaluated task and is not assumed to be universally appropriate for all applications.

The Anomaly Detection Engine evaluates each authenticated participant update relative to a robust reference constructed from the updates submitted during the same federated round. Let $\mathrm{w}^{(t)}$ denote the global model distributed at the beginning of round $t$, and let $\mathrm{w}_i^{(t+1)}$ denote the locally trained model submitted by participant $i$. The participant update, robust reference update, update distance, standardized deviation, and normalized anomaly-risk score are defined as:

$$\Delta_i^{(t)} = \mathbf{w}_i^{(t+1)} - \boldsymbol{w}^{(t)},$$

$$\tilde{\Delta}^{(t)} = coordmed_{j \in S_t}\left(\Delta_j^{(t)}\right),$$

$$d_i^{(t)} = \left\|\Delta_i^{(t)} - \tilde{\Delta}^{(t)}\right\|_2,$$

$$u_i^{(t)} = \frac{d_i^{(t)} - median_{j \in S_t}\left(d_j^{(t)}\right)}{1.4826\, MAD_{j \in S_t}\left(d_j^{(t)}\right) + \epsilon_A},$$

$$A_i^{(t)} = \sigma\left(u_i^{(t)}\right). \quad (4)$$

Here, $\mathcal{S}_t$ is the set of participants selected in round $t$, coordmed denotes the coordinate-wise median, $\|\cdot\|_2$ denotes the Euclidean norm, MAD denotes the median absolute deviation, $\varepsilon > 0$prevents division by zero, and $\sigma(x) = 1/(1 + e^{-x})$ is the logistic function. The resulting anomaly-risk score satisfies $A_i^{(t)} \in (0,1)$, with larger values indicating greater deviation from the robust round reference. The standardized threshold $\tau_u$ was set to the 95th percentile of standardized deviations observed during benign IID and non-IID calibration rounds. The corresponding anomaly-risk threshold was $\tau_A = \sigma(\tau_u) = 0.9499$. An update was flagged when $u_i^{(t)} \geq \tau_u$, equivalently when $A_i^{(t)} \geq \tau_A$. Both thresholds were fixed before final test evaluation.

A high anomaly score is treated as evidence that an update is unusual and is not interpreted as conclusive proof of malicious behaviour. Legitimate non-IID data, local class imbalance, optimization instability, or concept drift may also produce divergent updates. Consequently, an isolated anomaly flag results in monitoring and reduced aggregation influence rather than automatic exclusion. Quarantine is applied only when the anomaly persists for $r_A$ consecutive evaluated rounds or is accompanied by a mandatory identity, integrity, evidence, or policy failure. The value of $r_A$ used in the prototype is reported in Section 5.2, and the false-positive rate for benign non-IID participants is reported in the experimental results.

The Tamper-Evident Audit and Runtime Monitoring Layer records the complete governance and aggregation history for every evaluated participant update. The recorded fields include the participant identifier, federated round identifier, policy version, update hash, evidence-verification status, verified governance scores, anomaly-risk score, temporal trust transition, accept, down-weight, or quarantine decision, and the final normalized aggregation weight. Runtime alerts are derived from verified governance events. Invalid signatures, update-hash mismatches, stale or replayed evidence, incorrect policy versions, falsified claims, or anomalies persisting for $r_A = 2$ rounds generate critical alerts and quarantine. A demographic-parity difference above $\tau_D$, explanation consistency below $\tau_E$, or an isolated anomaly risk at or above $\tau_A$ generates a warning and down-weighting or monitoring. Each alert is specified to record the participant and round identifiers, triggering metric, observed value, applicable threshold, policy version, update hash, decision, and timestamp.

### 3.2 Governance-Aware Aggregation

Traditional Federated Averaging assigns participant influence primarily according to the number of local training samples [1]. This approach does not account for verified governance behaviour and may allow anomalous or non-compliant participants to influence the global model. GAFL therefore combines participant dataset size with the bounded temporal trust score defined in Equation (1). Let $S_t$denote the set of participants selected during round $t$, $n_i$ the number of local training samples held by participant $i$, and $z_i^{(t)} \in \{0,1\}$ the participant eligibility indicator. The eligibility indicator is one when the participant passes all mandatory checks and zero when quarantined. Let $N_t = \sum_{j \epsilon S_t} z_j^{(t)} n_j$ denote the total sample count of eligible participants. The non-negative raw contribution score is defined as:

$$N_t = \sum_{j \in S_t} z_j^{(t)} n_j,$$

where $N_t$ is the total number of samples held by eligible participants.

$$r_i^{(t)} = z_i^{(t)} \left(\frac{n_i}{N_t}\right)^{\gamma} \left[\epsilon + (1 - \varepsilon) T_i^{(t)}\right]^{\alpha},$$

The normalized aggregation weight is

$$\omega_i^{(t)} = \frac{r_i^{(t)}}{\sum_{j \in S_t} r_j^{(t)}},$$

And the updated global model is:

$$\boldsymbol{w}^{(t+1)} = \sum_{i \in S_t} \omega_i^{(t)} \boldsymbol{w}_i^{(t+1)}. \tag{5}$$

Here, $\alpha \geq 0$ controls the influence of verified temporal trust, $\gamma \geq 0$ controls the influence of participant dataset size, and $0 < \varepsilon < 1$ is a smoothing constant that prevents an accepted participant from receiving an exact zero contribution solely because of a low but non-zero trust score. The fairness, explainability, compliance, and anomaly trade-offs are governed by the coefficients $\lambda_C$, $\lambda_F$, $\lambda_E$, and $\lambda_A$ used in Equation (1). Since $z_i^{(t)}$, the dataset-size term, and the trust term are all non-negative, $r_i^{(t)} \geq 0$. Whenever at least one participant is eligible, the normalized weights satisfy:

$$\omega_i^{(t)} \geq 0, \qquad \sum_{i \in S_t} \omega_i^{(t)} = 1$$

The resulting global model is therefore a convex combination of eligible participant models and cannot reverse a participant's contribution through a negative weight. If no participant is eligible during a round, the aggregation step is not executed; the server retains $w^{(t)}$and escalates the round for governance review. When $\alpha = 0$ and $\gamma = 1$, the formulation reduces to dataset-size-based Federated Averaging for all eligible participants. Increasing $\alpha$ gives greater influence to verified temporal trust, while changing $\gamma$ controls the relative importance of dataset size. The prototype used the policy-defined values of $\alpha$, $\gamma$, $\epsilon$, and the governance coefficients reported in Table 1. The reported sensitivity analysis varied only the demographic-parity threshold $\tau_D$.

## 4 Threat Model and Governance Mechanisms

GAFL assumes a semi-trusted cross-silo environment in which participants retain local data but may submit manipulated updates, invalid evidence, biased models, or falsified governance claims. The Independent Governance Evidence Verifier, Governance Orchestration Layer, policy repository, governance-validation dataset, Aggregation Server, and audit layer are trusted; participant nodes are semi-trusted. Signatures, nonces, timestamps, round and policy identifiers, and update-hash binding protect evidence integrity and freshness.

The threat model covers five categories: malicious model-update manipulation; biased but non-malicious local data; mandatory governance non-compliance; anomalous but potentially benign training behaviour; and falsified fairness, explanation, or compliance evidence. Statistical deviation is not treated as proof of maliciousness because non-IID data, class imbalance, optimization instability, and distribution shift may also produce unusual updates. Mandatory identity, integrity, freshness, or evidence failures cause quarantine, whereas fairness violations or isolated anomalies normally cause monitoring or down-weighting. For every round, the verifier authenticates the update and independently recomputes required governance metrics. The control plane combines verified evidence, anomaly risk, and temporal trust to assign accept, down-weight, or quarantine, after which only eligible updates enter Equation (5). The tamper-evident audit layer records evidence, verification outcomes, trust transitions, decisions, and weights. GAFL does not guarantee identification of malicious intent or address compromised trusted components, strong collusion, unrepresentative validation data, or model-update privacy leakage.

## 5 Prototype Implementation and Experimental Methodology

A prototype of the proposed GAFL framework was implemented as a simulated cross-silo federated learning environment containing 10 participant nodes, an Independent Governance Evidence Verifier, a Governance Orchestration Layer, a Governance-Aware Aggregation Server, and Tamper-Evident Audit and Runtime Monitoring Layer.

Each participant retained its local training data and transmitted only its model update and signed preliminary governance-evidence bundle. The verifier authenticated each participant, bound the evidence to the corresponding model update, and independently recomputed selected governance measurements using a trusted governance-validation dataset. The Governance Orchestration Layer then evaluated verified policy-compliance, fairness, explainability, anomaly, and temporal trust information before assigning an accept, down-weight, or quarantine decision. Eligible participant updates were aggregated using the normalized governance-aware weighting mechanism defined in Equation (5). The experimental evaluation examined predictive performance, fairness, malicious-update detection, evidence-verification effectiveness, participant trust, policy enforcement, convergence, and governance overhead under honest, benign non-IID, biased, malicious-update, non-compliant, falsified-evidence, and mixed-threat conditions.

### 5.1 Dataset and Preprocessing

The Adult Income dataset was used for binary classification and fairness-sensitive federated learning [11]. The target indicates whether annual income exceeds USD 50,000, and sex was retained separately as the protected attribute. Records containing missing values were removed, categorical variables were one-hot encoded, and continuous features were standardized using parameters fitted only on the training data. The processed data were divided into mutually exclusive training, governance-validation, and test partitions using a 70:15:15 ratio. The training partition was distributed among participants, the governance-validation partition was used for metric recomputation and threshold calibration, and the test partition was reserved for final model evaluation.

### 5.2 Federated Learning Configuration

The simulation used 10 cross-silo participants, all selected in each of 25 communication rounds. Each participant trained a single-layer logistic-regression model for two local epochs using Adam, a batch size of 64, and a learning rate of 0.01. All methods used identical data partitions, model initialization, hyperparameters, and five random seeds. Approximately IID data were allocated equally using joint stratification over income label and sex. For the benign non-IID condition, the joint label–sex strata were distributed using a Dirichlet concentration parameter of 0.5 [12]. This condition isolated statistical heterogeneity from malicious behaviour.

The governance and aggregation coefficients were fixed before testing and are reported in Table 1. In operational deployments, these coefficients may instead be selected on governance-validation data using a constrained objective balancing predictive performance, fairness, detection effectiveness, false interventions, and latency. The anomaly, explanation-consistency, and evidence-verification thresholds were calibrated using governance-validation data and fixed before final testing. When coefficients are not policy-mandated, candidate non-negative vectors summing to one may be compared on governance-validation data using predictive performance, demographic disparity, threat-intervention recall, benign-intervention rate, and latency; the final test set must remain unused during selection.

**Table 1**. Experimental and governance configuration of the GAFL prototype.

| Parameter | Implemented Value |
|---|---|
| Dataset; target | Adult Income; income $> \$50,000$ |
| Protected attribute | Sex |
| Preprocessing | Row removal; one-hot encoding; standardization |
| Train/validation/test split | 70:15:15 |
| Participants; selection | 10; 100% per round |
| Training | 25 rounds; 2 local epochs |
| Optimizer settings | Adam; batch 64; learning rate 0.01 |
| Model | Single-layer logistic regression |
| IID allocation | Joint label–sex stratified allocation |
| Non-IID allocation | Joint label–sex Dirichlet, $\delta = 0.5$ |
| Trust parameters | $\rho = 0.70$, $\alpha = 1.00$, $\gamma = 1.00$ $\varepsilon = 0.05$ |
| Governance coefficients | $\lambda_C = 0.35$, $\lambda_F = 0.25$, $\lambda_E = 0.15$, $\lambda_A = 0.25$ |
| DPD thresholds | $\tau_D = 0.10$; $sensitivity = 0.15$ |
| Anomaly configuration | 95th percentile; $\tau_A = 0.9499$; $r_A = 2$ |
| Explainability | Closed-form linear SHAP; $\tau_E = 0.9592$ |
| Verification tolerance | $\tau_V = 0.0241$ |
| Reproducibility | 5 seeds: 11, 22, 33, 44, 55 |

### 5.3 Participant and Threat Scenarios

Seven scenarios were evaluated using the participant configurations in Table 2. The honest scenario used approximately IID partitions, while benign non-IID data were generated using joint label - sex Dirichlet allocation with $\delta = 0.5$. In the biased-data scenario, $client_00$ and $client_01$ each had 94 of 117 female positive-income records removed without label modification, reducing their partitions from 3,167 to 3,073 samples; no replacement occurred because all training records were already allocated. The malicious-update scenario used two sign-reversed updates scaled by three. In the non-compliant scenario, two participants changed the signed policy version from $GAFL - v1.0$ to $GAFL - v0.9$, producing policy-version and signature failures. In the falsified-evidence scenario, two participants increased reported fairness and explanation scores by 0.25, capped at 1.0. The mixed-threat scenario included one participant from each threat category and six honest participants. Honest heterogeneous updates flagged by the anomaly mechanism were counted as anomaly false positives, while fairness-policy interventions were reported separately. Mandatory structural or falsified-evidence failures caused quarantine; fairness violations and isolated anomalies caused down-weighting unless persistent.

**Table 2**. Participant configuration under the evaluated scenarios.

| Scenario | Participant Configuration |
|---|---|
| Honest | 10 honest; approximately IID |
| Benign non-IID | 10 honest; Dirichlet non-IID |
| Biased data | 2 biased participants; 8 honest participants |
| Malicious update | 2 malicious participants; 8 honest participants |
| Non-compliant | 2 invalid-signature; 8 honest participants |
| Falsified evidence | 2 evidence-falsifying participants; 8 honest participants |
| Mixed threat | 1 biased, 1 malicious, 1 non-compliant, 1 falsifying, 6 honest participants |

### 5.4 Governance Validation Workflow

Governance validation was performed before aggregation in every federated round. The server transmitted the current global model, round identifier, policy version, and validation thresholds to the selected participants. After local training, participant $i$ submitted its model update and a governance-evidence bundle containing the participant identifier, round identifier, policy version, SHA-256 update hash, reported fairness and explanation scores, timestamp, nonce, and Ed25519 digital signature [13]. Binding the evidence to the update hash prevented evidence generated for one model update from being reused for another. Client-generated governance metrics were treated as unverified claims. The verifier checked the Ed25519 signature, round identifier, policy version, timestamp, nonce, evidence completeness, and update-hash binding. It independently recomputed demographic-parity difference and linear-SHAP explanation consistency using the governance-validation data. Reported and verified numerical values differing by more than $\tau_V = 0.0241$ were classified as falsified evidence; invalid, stale, replayed, or falsified evidence triggered quarantine.

The Anomaly Detection Engine calculated each update's risk using Equation (4). Because benign non-IID data may also produce divergent updates, an anomaly flag was treated as a risk signal rather than proof of malicious behaviour and was evaluated alongside validation impact, update direction, temporal trust, and repeated behaviour. Using verified compliance, fairness, explainability, anomaly, and prior trust information, the Temporal Trust Engine updated $T_i^{(t)}$ using Equation (1).

The Decision Engine then assigned accept, down-weight, or quarantine; any mandatory identity, integrity, evidence, or policy failure resulted in quarantine for that round. Accepted and down-weighted updates received non-negative normalized weights using Equation (5), while quarantined updates had $z_i^{(t)} = 0$ and contributed nothing. If no update was eligible, the server retained the previous global model and escalated the round for governance review. The tamper-evident audit layer recorded participant and round identifiers, policy version, update hash, verification outcome, governance scores, anomaly risk, trust transition, decision, aggregation weight, and timestamp. Records were linked using SHA-256 hashes to provide tamper evidence.

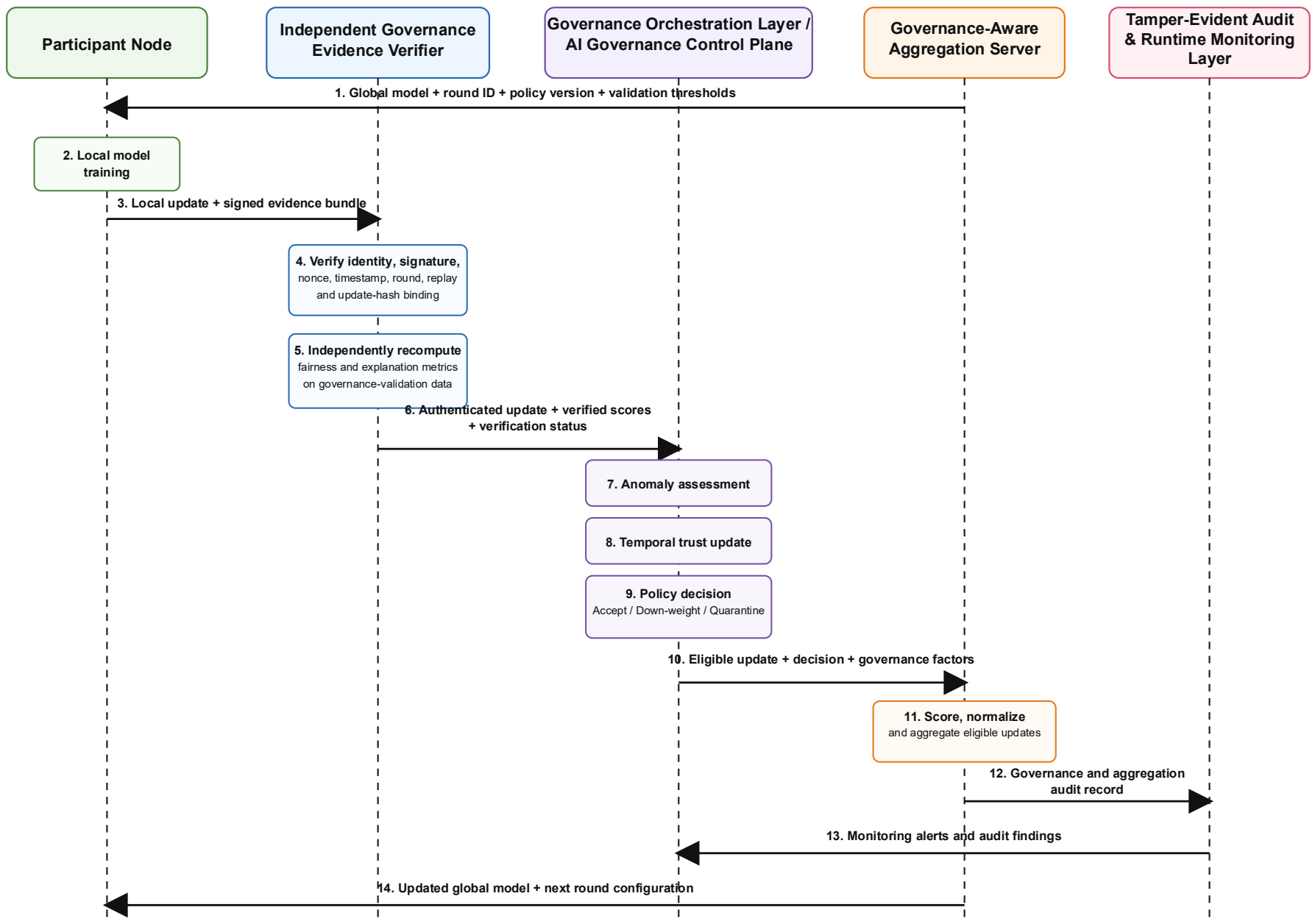

**Fig. 2.** Governance validation and aggregation sequence during one GAFL round.

Evaluation covered accuracy, macro-F1, convergence, demographic-parity difference, malicious-update intervention, benign non-IID anomaly false positives, policy and evidence-falsification detection, trust trajectories, quarantine correctness, and governance latency.

## 6 Results and Discussion

Table 3 compares FedAvg, anomaly-filtered FedAvg, GAFL without independent verification, and complete GAFL. Intervention rates represent affected participants assigned down-weighting or quarantine at the final round; latency is reported as mean ± standard deviation across five seeds.

**Table 3.** Final-round intervention rates and mean governance latency across five seeds.

| Method | Malicious | Non-compliant | Falsified Evidence | Mixed Threat | Latency (ms/round) |
|---|---|---|---|---|---|
| FedAvg | 0% | 0% | 0% | 0% | 3.82 ± 0.79 |
| FedAvg + anomaly filter | 100% | 0% | 0% | 25% | 4.05 ± 0.62 |
| GAFL-NoVerify | 100% | 100% | 0% | 75% | 4.45 ± 0.84 |
| GAFL-Full | 100% | 100% | 100% | 100% | 173.67 ± 26.94 |

Intervention rate denotes the proportion of affected participants assigned a down-weight or quarantine decision at the final round. FedAvg performs no governance checks and therefore has a zero intervention rate by construction.

Complete GAFL reduced the mean aggregation weight of malicious participants from 0.1000 under FedAvg to 0.0037, a 96.3% reduction. Non-compliant participants received zero aggregation weight, and all 250 non-compliant participant-round submissions were quarantined; each produced both a policy-version mismatch and a signature-invalidity finding. Under the malicious-update scenario, GAFL-Full achieved 0.8404 $\pm$ 0.0013 accuracy and 0.7733 $\pm$ 0.0032 macro-F1; under non-compliance, it achieved 0.8403 $\pm$ 0.0013 accuracy and 0.7731 $\pm$ 0.0033 macro-F1. Corresponding FedAvg results were 0.8407 $\pm$ 0.0006 accuracy and 0.7734 $\pm$ 0.0018 macro-F1 in both scenarios.

Across five seeds, mean final trust was 0.7407 $\pm$ 0.0088 for honest participants, 0.3466 $\pm$ 0.0068 for malicious participants, and 0.4790 $\pm$ 0.0014 for non-compliant participants, from an initial value of 0.50. Honest benign non-IID participants retained 0.6906 $\pm$ 0.0274 trust. At the final round, 15 of 50 honest non-IID observations were quarantined - four for persistent anomaly and eleven for evidence mismatch - while 26 were down-weighted for fairness. Increasing $\tau_D$ from 0.10 to 0.15 reduced all-round intervention rates only modestly: from 99.76% to 98.80% under IID data and from 76.32% to 74.64% under benign non-IID data. Using $\tau_A = 0.9499$, the anomaly mechanism flagged 100% of malicious updates and 9.68% of honest benign non-IID updates. Under the malicious-update scenario, GAFL-Full achieved 0.8404 $\pm$ 0.0013 accuracy and 0.7733 $\pm$ 0.0032 macro-F1, compared with 0.8407 $\pm$ 0.0006 and 0.7734 $\pm$ 0.0018 for FedAvg, and 0.8402 $\pm$ 0.0014 and 0.7731 $\pm$ 0.0034 for anomaly-filtered FedAvg. Thus, the implemented attack was detected without a material predictive-performance improvement over the baselines.

Under the biased-data scenario, GAFL-Full achieved a demographic-parity difference of 0.1810 $\pm$ 0.0059, compared with 0.1859 $\pm$ 0.0023 for FedAvg, 0.1827$\pm$ 0.0038 for FedAvg with anomaly filtering, and 0.1820 $\pm$ 0.0042 for GAFL-NoVerify. All 250 biased-participant-round observations exceeded the primary threshold $\tau_D = 0.10$. Their mean aggregation weight decreased from 0.0977 under FedAvg to 0.0839 under GAFL-Full, a 14.1% reduction, while accuracy remained effectively unchanged at 0.8408 for both methods. However, the honest-scenario disparity was also 0.1754 $\pm$ 0.0024, showing that the strict threshold surfaced a broader fairness limitation of the unmitigated base model rather than uniquely distinguishing biased participants. At the final round, the mean verified explanation-consistency score was 0.9987 $\pm$ 0.0001 for honest participants and 0.9530 $\pm$ 0.0037 for malicious-update participants. In the falsified-evidence scenario, GAFL-Full identified and quarantined all 250 falsified participant-round claims, whereas GAFL-NoVerify identified none because it used self-reported governance values without independent recomputation. This ablation demonstrates that independent verification was necessary to detect the implemented evidence-falsification attack.

GAFL-Full required 173.67 ± 26.94 ms of governance processing per round, compared with 3.82 ± 0.79 ms for FedAvg and 4.45 ± 0.84 ms for GAFL-NoVerify. This corresponds to approximately 45.4 times the FedAvg latency. Independent fairness and explanation recomputation required approximately 84.98 ms each and jointly accounted for 97.9% of GAFL-Full latency. All 140 primary-run audit hash chains passed integrity verification; however, audit-writing latency was not separately measured.

The findings are limited to the implemented cross-silo simulation, selected dataset, predictive model, threat configurations, and governance-validation partition. Results may differ with larger participant populations, stronger collusion, compromised trusted components, privacy attacks, unstructured data, or large-scale deep models. GAFL should therefore be interpreted as an experimentally evaluated governance architecture under the stated assumptions rather than a complete defence against all federated-learning threats. The biased-data intervention also reduced each affected partition by 3.0% because no unassigned replacement records were available; consequently, its measured effects cannot be attributed solely to demographic composition. The evaluation did not include direct experimental comparisons with FLTrust, q-FFL, Krum, or coordinate-wise trimmed mean. These methods address complementary trust bootstrapping, device-level fairness, or Byzantine-robust aggregation objectives, whereas the present experiments focused on isolating GAFL's governance components through FedAvg, anomaly-filtering, and evidence-verification ablations. Direct controlled comparisons with these methods remain necessary future work.

## 7 Conclusion and Future Work

This paper introduced GAFL, a federated-learning architecture integrating independent governance-evidence verification, temporal trust, fairness and explanation assessment, anomaly monitoring, policy enforcement, normalized aggregation, and tamper-evident auditing. In experiments with 10 participants and five seeds, GAFL-Full intervened on all affected participants at the final round and detected all 250 falsified participant-round claims, whereas GAFL-NoVerify detected none. Malicious-participant aggregation weight decreased by 96.3%, while predictive performance remained broadly similar to FedAvg. These governance benefits incurred substantial cost: GAFL-Full required 173.67 ± 26.94 ms per round, approximately 45.4 times the FedAvg latency. Moreover, frequent interventions for honest participants exposed sensitivity to the fairness threshold and verifier tolerance. Future work should combine governance enforcement with fairness-aware training, improve evidence-verification calibration, reduce explanation-computation overhead, and evaluate larger participant populations and stronger attacks.

**Acknowledgments.** The authors would like to acknowledge the guidance and support received during the development of this research work.

**Disclosure of Interests.** The authors declare that there are no financial or personal relationships that could have influenced the work reported in this paper.